\documentclass[final]{cvpr}
\usepackage{times}
\usepackage{epsfig}
\usepackage{graphicx}
\usepackage{amsmath}
\usepackage{amssymb}
\usepackage{algorithm}
\usepackage{algorithmicx}
\usepackage[noend]{algpseudocode}
\usepackage{multirow}
\usepackage{color}
\usepackage{makecell}
\usepackage[pagebackref=true,breaklinks=true,colorlinks,bookmarks=false]{hyperref}

\def\cvprPaperID{5806}
\def\confYear{CVPR 2021}

\newcommand{\cgreen}[1]{{\color{black} #1}}

\begin{document}

\title{ Combined Depth Space based Architecture Search For Person Re-identification  }
\author{Hanjun Li\textsuperscript{1,4},
Gaojie Wu\textsuperscript{1},
Wei-Shi Zheng\textsuperscript{1,2,3,}\thanks{Corresponding author}
\\
\textsuperscript{1}{School of Computer Science and Engineering, Sun Yat-sen University, China}\\
\textsuperscript{2}{Peng Cheng Laboratory, Shenzhen 518005, China}\\
\textsuperscript{3}{Key Laboratory of Machine Intelligence and Advanced Computing, Ministry of Education, China}\\
\textsuperscript{4}{Pazhou Lab, Guangzhou, China} \\
\tt\small lihj85@mail2.sysu.edu.cn, wugj7@mail2.sysu.edu.cn,wszheng@ieee.org
}
\maketitle
\thispagestyle{empty}

\begin{abstract}
\cgreen{Most works on person re-identification (ReID) take advantage of large backbone networks such as ResNet, which are designed for image classification instead of ReID, for feature extraction. However, these backbones may not be computationally efficient or the most suitable architectures for ReID. In this work, we aim to design a lightweight and suitable network for ReID. We propose a novel search space called Combined Depth Space (CDS), based on which we search for an efficient network architecture, which we call CDNet, via a differentiable architecture search algorithm. Through the use of the combined basic building blocks in CDS, CDNet tends to focus on combined pattern information that is typically found in images of pedestrians. We then propose a low-cost search strategy named the Top-k Sample Search strategy to make full use of the search space and avoid trapping in local optimal result. Furthermore, an effective Fine-grained Balance Neck (FBLNeck), which is removable at the inference time, is presented to balance the effects of triplet loss and softmax loss during the training process.} Extensive experiments show that our CDNet ($\sim$1.8 M parameters) has comparable performance with state-of-the-art lightweight networks. 
\end{abstract}

\section{Introduction}
\cgreen{Person re-identification (ReID) aims to retrieve images of a specific person from different surveillance cameras. Since AlexNet \cite{krizhevsky2017imagenet} was proposed in the ILSVRC-2012 \cite{deng2009imagenet}, convolution neural networks (CNNs) have become increasingly popular. With the emergence of complex models \cite{szegedy2015going,simonyan2014very,he2016deep}, people tend to utilize them as backbones to achieve higher performance on the ReID task. However, there are two obvious drawbacks to such implementations. First, they rely deeply on the performance of the backbone and limit researchers to explore more suitable network architectures for ReID. Second, these backbones require large computational resources and time costs at inference time, making them unaffordable for some practical/edge devices with limited computing resources, such as intelligent surveillance cameras. Instead, by deploying a lightweight network across a number of surveillance cameras, only the features these devices extract need to be gathered to retrieve the target person, which is much faster than gathering raw images and processing them with very large backbone networks. For these reasons, we aim to construct a lightweight network that is computationally efficient and more suitable for ReID.}
\begin{figure}[t]
\begin{center}
		\includegraphics[width = 0.9\linewidth]{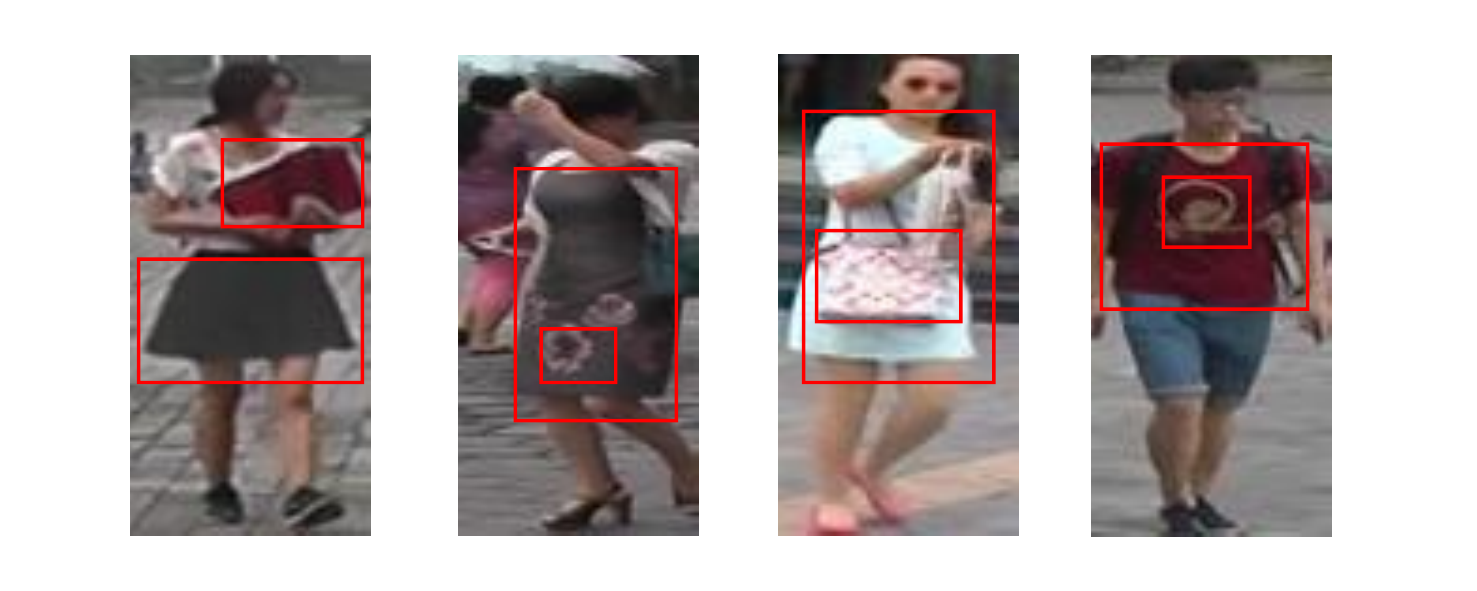}
\end{center}
\caption{Note that paired salient objects of different sizes can be commonly found in pedestrian images. Our CBlock is designed to explicitly learn such combined patterns.}
	\label{fig:pairs}
\end{figure}

In recent years, Neural Architecture Search (NAS) has been utilized to search lightweight but effective networks. \cite{zoph2016neural} takes 2000 GPU days to search the NASNet via reinforcement learning, which is far too long for most of researchers. To reduce the expensive search cost, \cite{liu2018darts} proposes a novel algorithm called differentiable architecture search (DARTS) using gradient descent, which dramatically reduces the search cost to 4 GPU days. Although the searched network architecture is very small, this method still has a few drawbacks. (1) The cell contains numerous complex connections, which are detrimental to parallel computing. \cite{lin2020neural} also notes that the irregular cell structures are not GPU friendly. (2) The algorithm only searches for normal and reduction cells and applies them to different layers. We argue that CNNs tend to concentrate on different pattern information at different depths and thus need to distinguish structures at different layers. (3) During the search, the algorithm computes each branch during forward propagation, even though the contributions of some branches with lower probabilities are negligible, resulting in heavy computational costs. As for the third drawback, \cite{dong2019searching} chooses to only compute the branch with the maximum weight between two nodes for forward propagation. However, this method may easily become trapped in a single local optimal network architecture since the gradient is mainly updated in the selected branch and other possible branches are gradually ignored; thus, the method cannot make full use of the search space. Apart from the above problems, we observe that most current search spaces are unable to explicitly learn combined pattern features for ReID (see Fig. \ref{fig:pairs}), which are known to have very strong discriminative value.

\cgreen{To address the aforementioned problems and make full use of the advantages of NAS to search lightweight networks, we propose a novel search space called Combined Depth Space (CDS) and a new search strategy called the Top-k Sample Search. In CDS, we design an efficient Combined Block (CBlock) consisting of two independent branches with different kernel sizes for explicitly learning combined pattern information. In this way, our CBlock only has two parallel branches and is thus GPU friendly. Moreover, our Top-k Sample Search computes the top-k branches according to the weights during the forward propagation, avoiding the computation of negligible branches or becoming trapped in a single local optimal network architecture, as is the case for \cite{dong2019searching}. In this way, we can not only largely reduce the search cost but also obtain a competitive lightweight network. Different from \cite{liu2018darts}, we choose to search the cells for each layer independently.

In addition, we jointly optimize the softmax loss and triplet loss for training, as in many works \cite{hermans2017defense,luo2019bag,quan2019auto}. Particularly, we further propose a simple but effective Balance Neck (BLNeck) to resolve the inconsistency between the targets of these two losses in the embedding space. In \cite{luo2019bag}, although BNNeck is presented to balance the effects of these losses, it does not always work for arbitrary network architectures (as shown in Table \ref{tb:fblneck}). However, the proposed BLNeck has a strong ability to map an embedding space constrained by the triplet loss to one constrained by the softmax loss; thus, the two losses can be optimized harmoniously. The stripe strategy is always used for extracting local features to guide the model to focus on more detailed information. We thus also integrate this idea into our Balance Neck, 
obtaining a new neck structure called the Fine-grained Balance Neck (FBLNeck) to further improve the performance.}

In summary, the contributions of this paper are summarized as follows:
\begin{itemize}
\item We propose a novel search space called Combined Depth Space (CDS), in which the CBlocks explicitly learn combined pattern features and are more suitable for ReID.
\item We propose a new search strategy called layer-wise Top-k Sample Search, which can largely reduce the search cost over that of other search strategies and make full use of the search space.
\item We propose a simple but effective Fine-grained Balance Neck (FBLNeck) for balancing the effects of triplet loss and softmax loss to better leverage their advantages.
\end{itemize}

The extensive experiments show that our CDNet achieves state-of-the-art performance on both ReID and other tasks among lightweight networks.
\section{Related Works}
\subsection{Lightweight Networks}
In recent years, to reduce the computational complexity of CNNs, some researchers have begun to explore effective and small-size models. MobileNets \cite{howard2017mobilenets,sandler2018mobilenetv2,howard2019searching} utilize depthwise separable convolution and reduce the number of parameters largely while maintaining comparable performance to standard networks. ShuffleNets \cite{zhang2018shufflenet,ma2018shufflenet} utilize group convolution to further reduce the number of parameters. Particularly, \cite{ma2018shufflenet} argues that an excessive number of branches will impede parallel computation, which we have also addressed in the present work. Other researchers obtain small-size networks via knowledge distillation \cite{wu2019distilled}, quantization \cite{han2016deep, wu2016quantized, jacob2018quantization}, network pruning \cite{han2016deep, li2016pruning, wang2020pruning, liu2018rethinking}, and so on. The greatest drawback of these design methods is that numerous experiments need to be conducted to empirically determine the best network structure.

To automate the architecture design process, reinforcement learning and evolution learning have been introduced to search  efficient network architectures with competitive accuracy on classification tasks \cite{zoph2016neural,real2019regularized,tan2019mnasnet}. After \cite{liu2018darts} proposed their differentiable architecture search algorithm via gradient decent, a number of researchers published extended works \cite{yang2020cars,dong2019searching,wu2019fbnet,chen2019renas} with similar algorithms. However, most of these methods adopt the search space and algorithm as DARTS, which has a number of drawbacks, as mentioned in the Introduction. Therefore, we propose a new search space called the Combined Depth Space, which is more efficient and suitable for the task of ReID.
\subsection{Person Re-Identification}
Recently, most of the proposed ReID models mainly utilize a complex network (\eg ResNet) as the backbone and integrate some special structures to extract extra information to enhance the discriminative features. \cite{sun2018beyond,zheng2019pyramidal,wang2018learning} utilize the stripe strategy to jointly extract both global features and local features and achieve great performance.
 \cite{zhang2019densely,song2018mask,wang2019spatial,zhu2019aware} introduce extra information such as body masks, camera information, and view labels to further improve the performance of the network. Noticeably, most of these methods adopt ResNet as the backbone, which has a large number of parameters and needs considerable computational resources. In practice, however, such a model is difficult to deploy in edge devices such as surveillance cameras with limited computational resources. We thus must avoid utilizing ResNet and aim to build an efficient but lightweight network to satisfy these specific demands.

Indeed, some works \cite{li2018harmonious,lawen2020compact,zhou2019omni} have been published describing the design of small networks for ReID. \cite{quan2019auto} introduces a part-aware block into the search space in DARTS and search a lightweight network for ReID.  \cite{zhou2019omni} proposes OSNet, which can learn omni-scale features and achieve promising results on both ReID and classification tasks. As a result of the special structure of OSNet, four branches compute the features in parallel at each layer; however, this leads to heavy computational resource consumption despite it only has 2.2 M parameters. We argue that an excessive number of parallel branches tend to extract redundant information; thus, we can discard specific branches to improve efficiency.
\section{Methodology}
In this section, we describe the proposed CDS in section \ref{sec:cds}. Then, the Top-k Sample Search algorithm is shown in section \ref{sec:tss}. Finally, we introduce the FBLNeck in section \ref{sec:fbl}.
\subsection{Combined Depth Space}
\label{sec:cds}
\begin{figure}[t]
\begin{center}
		\includegraphics[width=0.9\linewidth]{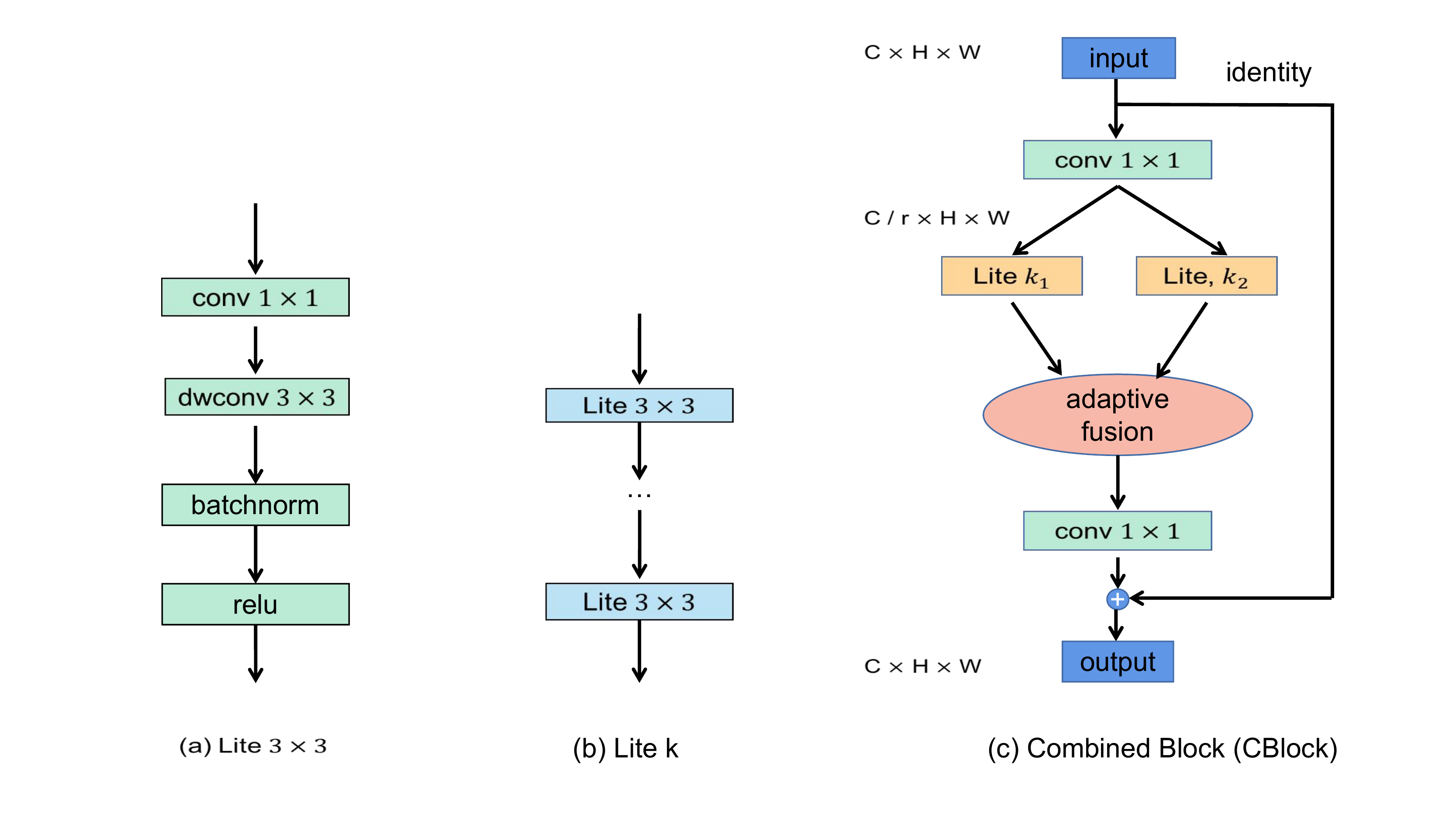}
\end{center}
\caption{(a) Lite $3\times3$ is the most basic building block. (b) Lite k consists of $\lfloor \frac{k}{2} \rfloor$ Lite $3\times3$. (c) CBlock combines two kinds of kernel $(k_1, k_2)$, and C $\times$ H $\times$ W denotes the current shape of the tensor.}
	\label{fig:cblock}
\end{figure}

To explore a more suitable architecture for ReID, we design a new efficient search space called Combined Depth Space (CDS), in which the building blocks explicitly learn combined pattern features. Based on CDS, the depth of the search network can adaptively vary within a specific range. Before defining the CDS, we first introduce the basic blocks. As shown in Fig. \ref{fig:cblock} (a), we adopt Lite 3$\times$3 from \cite{zhou2019omni} as our basic block.

\noindent \textbf{- CBlock.} As shown in Fig. \ref{fig:cblock}(c), our CBlock utilizes two different kernels with different receptive fields to jointly learn various scale patterns. We then elaborately fuse the features learned and obtain more discriminative features. Therefore, let $CK$ =\{(3,5), (3,7), (3,9), (5,7), (5,9), (7,9)\}; given the combination $(k_i, k_j)\in CK$, we can obtain 6 types of CBlock. To reduce the number of parameters and computation costs, we replace the depthwise convolution k $\times$ k with $\lfloor \frac{k}{2} \rfloor$ Lite 3$\times$3 (see in Fig. \ref{fig:cblock}(b)) since their final feature maps share the same receptive field, and there are fewer parameters with the latter than with the former. Let $g$ denote Lite 3$\times$3. Given input $x$ and kernel size k, we can formulate the computation process of Lite $k$ as follows:
\begin{equation}
	Lite(x, k) = g(x) \circ \lfloor \frac{k}{2} \rfloor
\end{equation}
op $\circ$ t means that op is run t times consecutively. Because the features learned from the two kernels are heterogeneous, it would be improper to simply sum them. As \cite{zhou2019omni} introduces an Adaptive Fusion Gate, we compute the weights of each channel according to the input and then compute the channel-wise weighted summation of $ Lite(x, k_1)$ and $Lite(x, k_2)$. The two conv 1$\times$1 are utilized to squeeze and restore the channels at a ratio r. 

\begin{figure}[t]
\begin{center}
		\includegraphics[width=0.9\linewidth]{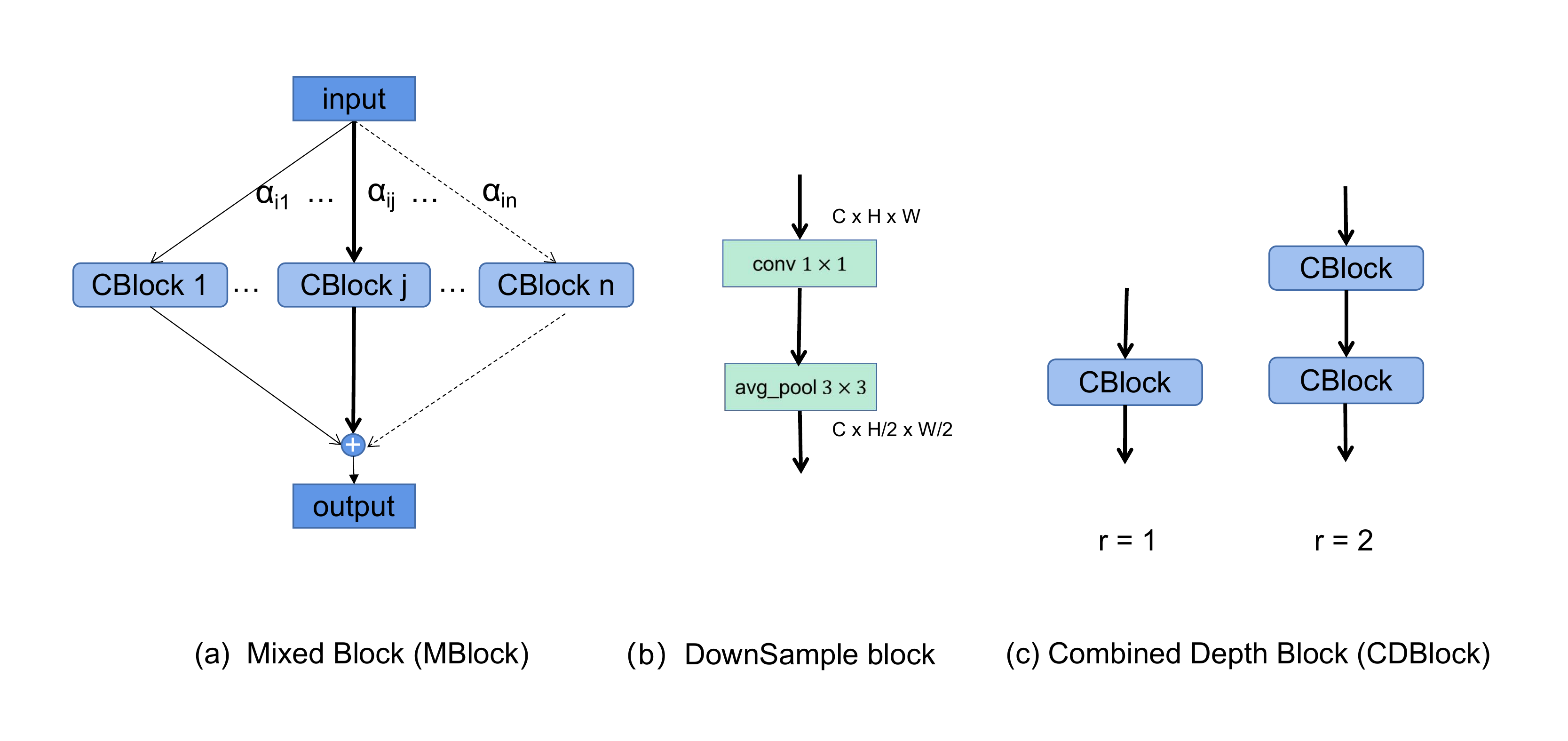}
\end{center}
\caption{MBlock consists of different CBlocks. Each branch has its own architecture parameters $\alpha_{ij}$. (b) A DownSample block is utilized to reduce the spatial size of the feature map between any two stages. (c) CDBlock consists of one or two CBlocks.}

	\label{fig:mblock}
\end{figure}

\noindent \textbf{- MBlock.} Our MBlock is equivalent to the cell in DARTS. However, we simplify the search cell and only need to select the appropriate combination of kernels rather than determine the complex connections between the inner nodes. Specifically, here, the combination of CBlocks belongs to the set of $CK$, that is, MBlock only has six candidate operations. As shown in Fig. \ref{fig:mblock}(a), each branch has its own weight $\alpha_{ij}$, representing the importance of the current CBlock $j$.

\begin{figure*}
\begin{center}
		\includegraphics[width=\linewidth]{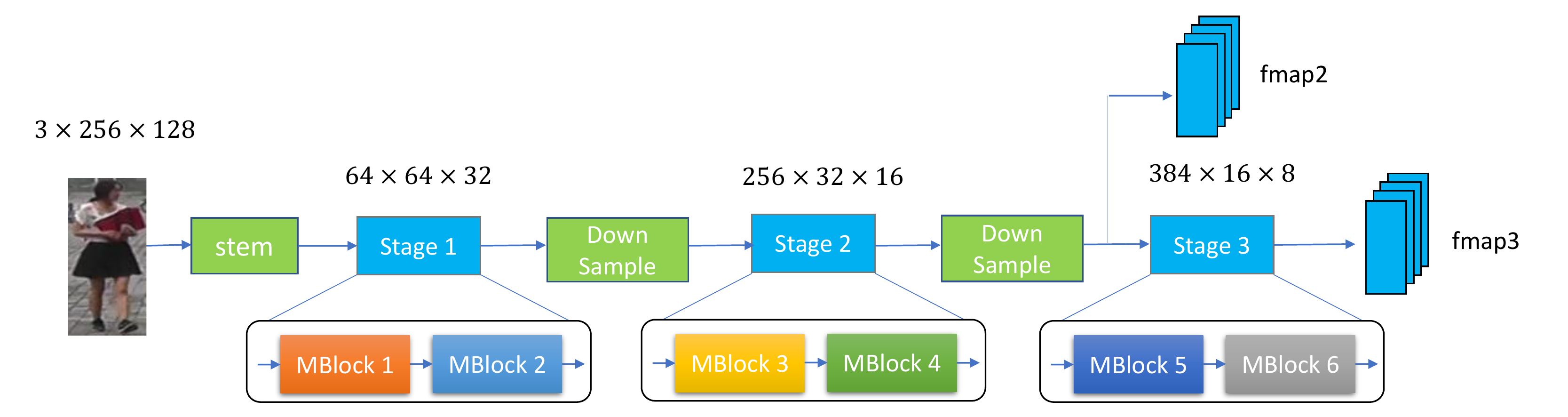}
\end{center}
\caption{The macro architecture of the search network for person re-identification. It includes a stem and three stages in total. Each stage stacks two MBlocks. Between every two stages, we insert a DownSample block to reduce the spatial size. The description c $\times$ h $\times$ w over each stage denotes the number of channels and the height and width of the in-tensor.}
	\label{fig:framework}
\end{figure*}

DARTS only searches one normal cell and one reduction cell and shares the inner structure across different layers. We argue that blocks at different depths of the neural network may focus on different information; thus, it would be sub-optimal to simply search two cells and stack the searched cells until comparable performance is obtained. Because of our efficient search strategy (described in section \ref{sec:tss}), we design a macro framework first, as shown in Fig. \ref{fig:framework}, and search independent cells for each layer during the neural architecture search. At the beginning of the search network, we adopt a normal convolution stem consisting of a 7$\times$7 standard convolution with stride 2 and 3$\times$3 max pooling with stride 2. As we can see, there are 6 MBlocks to be searched at different depths. Unlike DARTS, we utilize a fixed reduction block, the DownSample block (Fig. \ref{fig:mblock}(b)).

Based on the aforementioned search space, termed Combined Space (CS), we call the searched architecture network CNet. Since each MBlock has 6 candidate operations, we can easily determine that the total size of the CS is $6^6\approx10^{4.7}$. Compared with the space of the normal NAS methods, the size of our CS is relatively small; nevertheless, it contains a sufficient number of efficient and high-quality structures.

\noindent \textbf{- CDBlock.} Note that our search network only has 6 MBlocks, which is relatively shallower than ResNet with its 16 building blocks. We can simply add one or two MBlocks to each stage to deepen our network, but this will greatly increase the number of parameters and computations. To effectively deepen the network, an appropriate number of blocks should be allocated at each stage rather than randomly or uniformly. To achieve this, we introduce a depth factor into CS, forming a new search space termed Combined Depth Space (CDS). We redefine $CK$ as $CDK$ = \{(3,5,1), (3,7,1), (3,9,1), (5,7,1), (7,9,1), (3,5,2), (3,7,2), (3,9,2), (5,7,2), (7,9,2)\}; let tuple $ (k_1, k_2, r) \in CDK$. $k_1, k_2$ denote the kernel size for two branches in CBlock. $r$ denotes the number of times CBlock is repeated. Given the redefined $CDK$, we can construct CDBlock as shown in Fig. \ref{fig:mblock}(c). Therefore, we can replace the CBlock in the original MBlock with CDBlock to construct a new MBlock. Similarly, we can easily compute the size of the CDS as $12^6 \approx 10^{6.5}$. After considering the depth factor, the depth of our network can range from 6 to 12, making our network more flexible. With the CDS, we can adaptively search for the optimal network depth rather than determine the network depth manually.

\subsection{Top-k Sample Search}
\label{sec:tss}
We denote the architecture parameters as $\alpha$ = \{$\alpha_1, \alpha_2, ... , \alpha_6$ \}, where each $\alpha_i$ is a $n$-dim vector that denotes the weights of each branch in the $i_{th}$ MBlock. Specially, $n$ is 6 for the CS and 12 for the CDS. We denote the network parameters of the search network as $w$. Therefore, our goal is to find the optimal architecture parameters $\alpha^*$ that can achieve a minimum validation loss $L_V$ after $w$ is updated by minimizing the training loss $L_T$, as shown in Eq.\ref{eq:object}.
\begin{equation}
\begin{aligned}
min_\alpha \quad &L_{V}(w^*(\alpha), \alpha) \\
&s.t. \quad w^*(\alpha) = argmin_w L_{T}(w,\alpha)
\end{aligned}
\label{eq:object}
\end{equation}

Obviously, the first impulse is to solve the top equation above by updating $w$ until it converges on the training set and then updating $\alpha$ until it converges on the validation set. These two steps are repeated until the search network reaches a state of convergence. However, attempting to achieve convergence in each iteration is time consuming, making this method unsuitable for finding the optimal result. Similar to \cite{dong2019searching, wu2019fbnet}, $w$ is instead updated using single batch training data, and $\alpha$ is updated using single batch validation data at each epoch. After updating $w$ and $\alpha$ alternatively for an adequate number of epochs, the optimal value $\alpha^*$ for Eq. \ref{eq:object} is ultimately approximated.

During the search, we only compute the top-k branches according to $\alpha_i$ for MBlock $i$. Given input $x$ and MBlock $i$, we denote the forward propagation as $F(i, x)$ and the $j_{th}$ branch in MBlock as $f_j$. Let $n$ denote the number of branches; the probability distribution of MBlock $i$, $p_i = [p_{i1},p_{i2},..., p_{in}]$, is obtained from Eq.\ref{eq:prob},
\begin{equation}
	p_{ij} = \cfrac{exp(\alpha_{ij})}{\sum_{j=1}^{n}exp(\alpha_{ij})}
	\label{eq:prob}
\end{equation}
We then construct a binary vector $h_i = [h_{i1}, h_{i2}, ... ,h_{in}]$ via Eq. \ref{eq:p2h}, where $h_{ij}$ is 1 if the $j_{th}$ branch is selected and 0 if it is dropped in forward propagation.
\begin{equation}
	h_{ij} = \begin{cases}
		 1,\quad p_{ij} \geq v_k \\
		 0, \quad p_{ij} < v_k 
	\end{cases}
	\label{eq:p2h}
\end{equation}
where $v_k$ denotes the $k_{th}$ largest value in $p_i$. Subsequently, we can formulate the forward propagation of MBlock $i$ as follows:
\begin{equation}
	F(i, x) = \sum_{j=1}^{n} h_{ij}f_j(x)
\label{eq:forward}
\end{equation}
However, since $h_i$ is a discrete distribution from a probability, $F(i, x)$ cannot back propagate gradients to $\alpha_i$ via $h_i$. To allow back propagation, we aim to build a bridge between $h$ and $\alpha$.
\begin{equation}
	m_i = h_i^* - p_i^* 
\end{equation}
\begin{equation}
	\hat{h_i} = m_i + p_i 
	\label{eq:newh}
\end{equation}
where $p_i^*, h_i^*$ denotes a copy of $p_i,h_i$ without gradients. Thus $m_i$ is a normal vector without gradients. We then replace $h_{ij}$ in Eq. \ref{eq:forward} with $\hat{h_{ij}}$. Since the gradients of $p_i$ need to be computed, the gradients of $\hat{h_i}$ in Eq.\ref{eq:newh} also need to be computed. In this way, we can compute the gradient of $\alpha_{ij}$ in MBlock $i$ as follows:
\begin{equation}
	\cfrac{\partial F(i,x)}{\partial \alpha_{ij}} = \sum_{k \in topk}(\cfrac{\partial F(i,x)}{\partial \hat{h_{ik}}}\cfrac{\partial \hat{h_{ik}}}{\partial p_{ik}}\cfrac{\partial p_{ik}}{\partial \alpha_{ij}})
	\label{eq:partial}
\end{equation}
The Eq. \ref{eq:forward} and Eq. \ref{eq:partial} show that we treat the top-k branches equally during forward propagation (each weight is 1) but update the architecture parameters according to its real weight $p$. Our search algorithm can be summarized as Algorithm \ref{alg:topk}.
\begin{figure}[t]
\begin{center}
		\includegraphics[width=\linewidth]{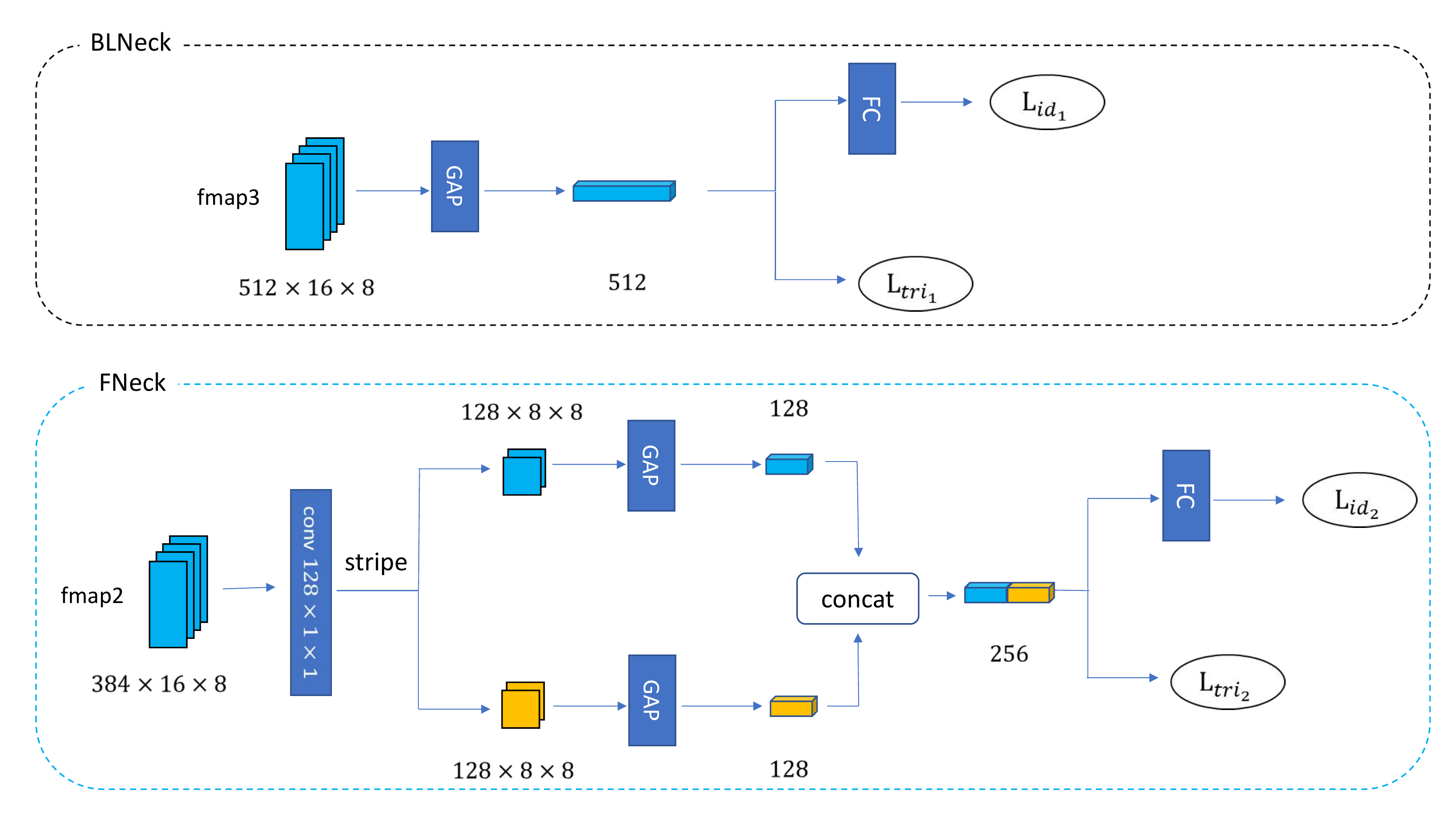}
\end{center}
\caption{The schematic of the Fine-grained Balance Neck consisting of BLNeck and FNeck. fmap2 and fmap3 are shown in Fig. \ref{fig:framework}. The main idea is to insert a fully connected layer between feats, which is constrained by the triplet loss, and feats, which is constrained by the softmax loss. Before partitioning fmap2, we first squeeze the channel to 128-dim.}
	\label{fig:fblneck}
\end{figure}
\subsection{Fine-grained Balance Neck}
\label{sec:fbl}
To efficiently tackle the problem that the targets of the triplet and softmax losses are inconsistent in the embedding space, we propose the effective Fine-grained Balance Neck (FBLNeck). As shown in Fig. \ref{fig:fblneck}, the FBLNeck consists of two parts. The upper part is named Balance Neck (BLNeck), which only utilizes global information, and the lower part is named Fine-grained Neck (FNeck), which partitions fmap2 into two stripes and extracts local information. Note that FNeck also has similar structure as BLNeck for aforementioned losses. It is worth noting that our local features are extracted at a relatively shallow layer rather than at the last layer selected by most works. This is because the shallow layers have a smaller receptive field than the deeper layers and thus tend to focus on fine-grained features. Additionally, we desire to avoid using the same global features as above and instead make full use of earlier features, which would lead to greater discriminability. Importantly, the fc layer  mainly transforms the triplet-friendly embedding space into a softmax-friendly embedding space in both BLNeck and FNeck.
\begin{algorithm}[ht]
\caption{Top-k Sample Search}
	\label{alg:topk}
\begin{algorithmic}[1]
\Require
Split the training set into two disjoint sets: $D_{train}$ and $D_{val}$;
Given the total search epoch $E$, architecture parameter learning rate $\eta_\alpha$, network parameter learning rate $\eta_w$,
initialize $\alpha$ and $w$.
\Ensure
Derived the final architecture from the learned $\alpha$.
\For{$e=1$ to $E$}
\For {batch data $(X_t, Y_t)$ in $D_{train}$}
\State Compute the training loss $L_T(X_t, Y_t)$
\State Update $w$: $w^* = w - \eta_w \bigtriangledown_wL_T$
\State Sample batch data $(X_v, Y_v)$ from $D_{val}$
\State Compute the validation loss $L_V(X_v, Y_v)$
\State Update $\alpha$: $\alpha^* = \alpha - \eta_\alpha \bigtriangledown_\alpha L_V$
\EndFor
\EndFor
\end{algorithmic}
\end{algorithm}
During the search training, we do not utilize FNeck; thus, our search loss can be formulated as follows:
\begin{equation}
	L_{search} = L_{tri1} + L_{id1}
\end{equation}
During the training of CNet or CDNet, we calculate the objective loss as Eq. \ref{eq:Ltrain},
\begin{equation}
	L_{train} = L_{tri1} + L_{id1} + L_{tri2} + L_{id2}
	\label{eq:Ltrain}
\end{equation}
\section{Experiments}
\subsection{Search for CNet and CDNet}
\begin{table}[ht]
	\begin{center}
		\begin{tabular}{c|c|c|c|c|c}
			\hline 
			 Model & Top-k & Param & Time & rank-1 & mAP \\ 
			\hline
			\multirow{4}{*}{CNet} & top-1 & 1.4M & 30.4ms & 93.3 & 82.6 \\
			& top-2 & 1.4M & 40.0ms & \textbf{93.6} & \textbf{83.5} \\
			& top-3 & 1.3M & 51.2ms & 93.1 & 81.9 \\
			& top-4 & 1.4M & 57.7ms & 92.9 & 82.3 \\ 
			\hline 
			\multirow{4}{*}{CDNet}& top-1 & 2.1M & 27.3ms & 93.4 & 83.2 \\
			& top-2 & 1.8M & 25.3ms & \textbf{93.7} & \textbf{83.7} \\
			&top-3 & 1.9M &  48.4ms & 93.6 & 83.5 \\
			&top-4 & 1.6M & 43.1ms & 93.2 & 83.0 \\
			\hline 
		\end{tabular}
	\end{center}
	\caption{The performance of different searched architectures evaluated on Market1501. The time is the average time required to process one image during the search. M:Million}
	\label{tb:topk}
\end{table}
Different from normal NAS, we search the network architectures on the Market1501 \cite{zheng2015scalable}. According to Algorithm \ref{alg:topk}, we further split the training set of Market1501 into a new training set and a validation set. Additional details on the data preparation and experimental configuration are described in the supplementary materials. Considering the computational costs involved, we conduct a series of Top-k Sample Search, $k \in$\{1,2,3,4\}. After the search, we derive the architecture according to the learned $\alpha$. For each MBlock, we select $j_{th}$ CBlock or CDBlock with the maximum weight eventually. Based on CS and CDS, the architectures derived from the search network are called CNet and CDNet, respectively. We then evaluate the searched architectures on Market1501. As shown in Table \ref{tb:topk}, the top-2 sample search yields the best architecture for both CNet and CDNet. Top-1 sample search tends to become trapped in a local architecture and cannot produce the optimal architecture. As the number of computational branches increases, the performance of the searched architectures worsens because the excess branches may exhibit a competitive relationship. Specially, we compute all branches of the search network like DARTS and the time cost as in Table \ref{tb:topk} is 100ms and 83.4ms for CNet and CDNet respectively. Taking search cost and performance into account, we suggest that the top-2 sample search is the best choice. The speed of the top-2 sample search is 2.5$\times$ greater than that of DARTS. We show the final architectures of CNet and CDNet via the top-2 sample search in Table \ref{tb:arch}.
\begin{table}[ht]
	\begin{center}
		\begin{tabular}{c|c|c}
			\hline
			\multirow{2}{*}{Layer} & CNet & CDNet \\
			\cline{2-3}
			& $(k_1,k_2)$ & $(k_1,k_2,r)$ \\ 
			\hline
			1 & (5,7) & (3,5,1) \\
			2 & (7,9) & (3,7,2) \\
			3 & (7,9) & (5,7,2) \\
			4 & (7,9) & (5,9,1) \\
			5 & (7,9) & (5,7,2) \\ 
			6 & (3,5) & (5,7,1) \\	
			\hline
		\end{tabular}
	\end{center}
	\caption{The architectures of CNet and CDNet. $k_1,k_2$ denotes the kernel size of the two branches in CBlock, and $r$ denotes the number of times CBlock is repeated within CDBlock.}
	\label{tb:arch}
\end{table}
\subsection{Evaluation on Person Re-Identification}
\label{sec:reid}
\paragraph{Datasets and implementation details} We conduct experiments on three widely used person ReID datasets: Market1501 \cite{zheng2015scalable}, DukeMTMC \cite{ristani2016performance} and MSMT17 \cite{wei2018person}. For all experiments, all images are resized to a resolution of 256 $\times$ 128, and Random Erasing Augment in \cite{luo2019bag} is utilized to imitate occlusion. During the training of CDNet and CNet, we adopt a training scheme similar to that in \cite{zhou2019omni}. We set the number of stripes at FNeck to 2. Finally, all models are trained with triplet loss and softmax loss unless stated otherwise.
\begin{table*}[ht]
	\begin{center}
		\begin{tabular}{l|c|c|c|c|c|c|c}
			\hline
			\multirow{2}{*}{Model} & \multirow{2}{*}{Param(M)} & \multicolumn{2}{c|}{Market1501}& \multicolumn{2}{c|}{DukeMTMC} & \multicolumn{2}{c}{MSMT17} \\
			\cline{3-8}
			& & rank-1 & mAP & rank-1 & mAP & rank-1 & mAP\\
			\hline
			ShuffleNet$^*$\cite{zhang2018shufflenet} & $\sim$1.9  & 84.8 & 65.0 & 71.6& 49.9 & 41.5 & 19.9 \\
			MobileNet$^*$\cite{sandler2018mobilenetv2} & $\sim$2.14 & 87.0 & 69.5 & 75.2 & 55.8 & 50.9 & 27.0 \\
			OSNet\cite{zhou2019omni} & 2.2  & \textcolor{blue}{93.6} & 81.0 & 84.7 & 68.6 & 71.0 & 43.3\\ 
			HA-CNN\cite{li2018harmonious} & 2.7 & 91.2 & 75.7 & 80.5 & 63.8 & - & - \\
			Auto-ReID\cite{quan2019auto} & 11.4 &  89.7 & 72.7 & - & - & - & - \\
			BagofTrick$^+$\cite{luo2019bag} & $\sim$26 &  83.7 & 65.8 & 76.0 & 62.2 & - & - \\
			\textbf{CNet(ours)} & 1.44 &  \textcolor{blue}{93.6} & \textcolor{blue}{83.5} & \textcolor{blue}{86.0}& \textcolor{blue}{73.2} & \textcolor{blue}{73.3} & \textcolor{blue}{47.7} \\
			\textbf{CDNet(ours)} & 1.8 & \textcolor{red}{93.7} & \textcolor{red}{83.7} & \textcolor{red}{86.7} & \textcolor{red}{73.9} & \textcolor{red}{73.7} & \textcolor{red}{48.5} \\
			\hline   
			\hline 
			PCB(+RPP)\cite{sun2018beyond} & $\sim$26 & 93.8 & 81.6 & 83.3 & 69.2 & 68.2 & 40.4 \\
			VPM\cite{sun2019perceive} & $\sim$26 &  93.0 & 80.8 & 83.6 & 72.6 & - & - \\
			BagofTricks\cite{luo2019bag}& $\sim$26 &  94.5 & 85.9 & 86.4 & 76.4 & - & - \\
			IANet\cite{hou2019interaction}& $\sim$26 & 94.4 & 83.1 & - & - & 75.5 & 46.8 \\
			CtF\cite{wang2020faster} & $\sim$26 &  93.7 & 84.9 & 87.6 & 74.8 & - & -\\
			SCSN\cite{chen2020salience} & $\sim$26 & \textcolor{red}{95.7} & \textcolor{red}{88.5} & \textcolor{red}{90.1} & \textcolor{red}{79.0} & \textcolor{red}{83.0} & \textcolor{red}{58.0} \\ 
			OSNet\cite{zhou2019omni}& 2.2 &  94.8 & 84.9 & \textcolor{blue}{88.6} & 73.5 & 78.7 & 52.9 \\
			Auto-ReID\cite{quan2019auto} & 11.4 &  94.5 & 85.1 & 88.5 & 75.1 & - & -\\
			\textbf{CDNet(ours)} & 1.8 &  \textcolor{blue}{95.1} & \textcolor{blue}{86.0} & \textcolor{blue}{88.6} & \textcolor{blue}{76.8} & \textcolor{blue}{78.9} & \textcolor{blue}{54.7} \\ 
			\hline	
		\end{tabular}
	\end{center}
	\caption{Performance on the Market1501, DukeMTMC and MSMT17 datasets. All of the models listed in the top half of the table are trained from scratch, and those in the bottom half are pretrained on ImageNet. * denotes that the result comes from \cite{zhou2019omni}. + denotes that we reproduce the result. - denotes that the result is unavailable. The number of parameters is counted at inference time. Best and second best results are colored with red and blue respectively. It is clear that our models surpass most published models by a clear margin with the fewest number of parameters.}
	\label{tb:reid}
\end{table*}
\cgreen{
\paragraph{Result analysis} As shown in Table \ref{tb:reid}, among the models that are not pretrained with ImageNet, both CNet and CDNet achieve competitive performances.
 The margins for mAP are even larger. Compared with that of Auto-ReID, which incorporates a part-aware super block into the search space of DARTS and searches for an architecture similar to DARTS, the rank-1 accuracy and mAP of CNet are better by 3.9\% and 10.8\%, respectively, on Market1501 with $\sim$5$\times$ fewer parameters. Hence, these results indirectly demonstrate that our search spaces (CS, CDS) are superior to the search space in DARTS. Furthermore, although we utilize the Lite 3$\times$3 basic block proposed in OSNet, our CNet outperforms OSNet in terms of mAP by a large margin on both Market1501 and DukeMTMC with much fewer parameters. CNet can be interpreted as a subnet of OSNet, which verifies that many branches of OSNet are redundant and useless. In contrast, CNet is more effective and computationally economical without redundant branches. Although HA-CNN utilizes an attention mechanism and BagofTricks utilizes many training tricks, they are still inferior to our models without ImageNet pretraining. CDNet is the enhanced version of CNet with an increased adaptive depth, which further improves the performance with only a small increase in the number of parameters. Since pretraining on ImageNet yields better performance for most networks, we also pretrained CDNet on ImageNet. As shown in the bottom half of Table \ref{tb:reid}, CDNet still achieves competitive performance among the models listed on all datasets with only 1.8 M parameters. Compared with OSNet, CDNet achieves higher mAP on all datasets, suggesting that our model is more robust in identifying difficult positive identities. 
}
\subsection{Visualization of Combined Pattern Learning} 
\begin{figure}[h]
\begin{center}
		\includegraphics[width=\linewidth]{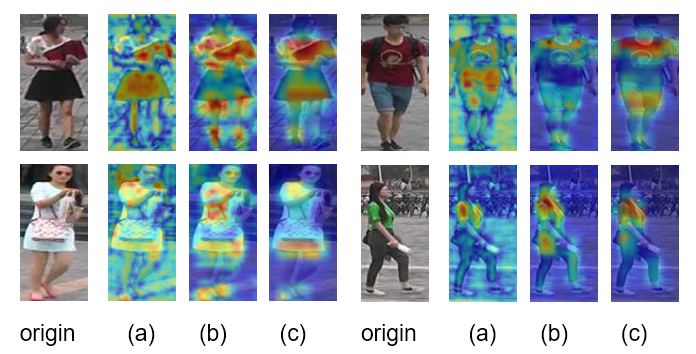}
\end{center}
\caption{Visualization of the learning representation from CDNet. The column origin corresponds to the original image. Columns (a), (b), and (c) correspond to the last feature maps of stage1, stage2 and stage3, respectively.}
	\label{fig:heatmap}
\end{figure}

CBlock is the core building block of CDNet and is designed to explicitly learn combined patterns in images. To verify that the proposed CDNet can learn discriminative combined features, we visualize the last feature maps at stage1, stage2 and stage3. The top left images in Fig. \ref{fig:heatmap} show a girl dressed in a black skirt with a red book in her hand. CDNet primarily captures the combined information on the skirt and the book. For the person in the bottom left images, CDNet mainly focuses on the handbag. Other than this salient object combination, plain patterns on clothing can also be captured by CDNet, as shown for the person in the top right images. Moreover, it can be observed that the legs of all identified persons are captured by CDNet, since the legs and the background can also be seen as a salient combination. Obviously, CDNet can effectively capture dominating information and ignore useless background information.
\subsection{Evaluation on Classification Task}
The CIFAR-100 \cite{krizhevsky2009learning} consists of 50K training images and 10K test images comprising 100 categories. The size of each image is 32 $\times$ 32. Random horizontal flip and random crop are utilized for data augmentation. We normalize the pixel values as in \cite{dong2019searching}, and the other training settings are the same as in section \ref{sec:reid}. As shown in Table \ref{tb:cifar100}, although GDAS and DARTS are originally searched for classification, they do not have significant advantages on CIFAR-100. With the exception of DARTS, our CDNet outperforms the other lightweight networks by a clear margin. In particular, CDNet is better than OSNet designed for ReID by 1.83\%. Obviously, the superior performance on classification task demonstrates the benefit of learning combined pattern information.
\begin{table}[h] 
	\begin{center}
		\begin{tabular}{l|c|c}
			\hline
			Model & Param(M) & Error(\%) \\
			\hline 
			GDAS\cite{dong2019searching} & 2.5 & 18.13 \\
			DARTS$^*$\cite{liu2018darts} & 3.4 & \textcolor{red}{17.54} \\
			DensNet\cite{huang2017densely} & 7.0 & 20.20 \\
			pre-act ResNet\cite{he2016identity} & 10.2 & 22.71 \\
			Wide ResNet\cite{zagoruyko2016wide} & 11.0 & 22.07 \\
			OSNet\cite{zhou2019omni} & 2.2 & 19.21 \\
			CDNet(ours) & 2.3 & \textcolor{blue}{17.83} \\
			\hline
		\end{tabular}
	\end{center}
	\caption{Error rates on CIFAR-100. * indicates that the result for DARTS is obtained from GDAS.}
	\label{tb:cifar100}
\end{table}
\subsection{Ablation Study}
\paragraph{- Best partitions for FNeck .} Related works \cite{sun2018beyond,zheng2019pyramidal,wang2018learning} split the final feature map to extract local features and subsequently enhance the discriminability by combining them with global features.
 However, they all only utilize the last feature map; thus, there is much similar information between global features and the local features, actually does little to improve discriminability.
 We propose exploring the use of shallower information for two reasons. First, it can avoid using the same information as global features, instead making full use of other features. Second, the shallower layer has a relatively small receptive field; thus, the local features are more fine-grained. In Table \ref{tb:partition}, we explore the suitable number of partitions for FNeck. By comparing the first two rows, we see that the features of the shallower layer indeed greatly contribute to improving performance. Both rank-1 accuracy and mAP are improved with a small increase in the feature-dim when the number of partitions increases to 2. As the number of partitions further increases, the mAP does not notably improve. Although CDNet achieves the best rank-1 accuracy (94.2\%) for 4 partitions, this would require considerable computational resources. Therefore, we choose 2 partitions for all experiments without further comment.
\begin{table}[ht] 
	\begin{center}
		\begin{tabular}{c|c|c|c}
			\hline
			Partition & Feature-dim & rank-1 & mAP \\
			\hline
			0 & 512 & 93.1 & 81.5 \\
			1 & 640 & 93.0 & 82.6 \\
			2 & 768 & 93.7 & \textbf{83.7} \\
			3 & 896 & 93.2 & 82.7 \\
			4 & 1024 & \textbf{94.2} & 83.1 \\
			\hline
		\end{tabular}
	\end{center}
	\caption{Effect of the number of FNeck partitions. The experiments are conducted with CDNet on Market1501. The entry for 0 partitions refers to an implementation without FNeck.}
	\label{tb:partition}
\end{table}
\vspace{-0.6cm}
\cgreen{
\paragraph{-Effect of FBLNeck.} To effectively make full use of the combination of the triplet loss and softmax loss, we propose a simple but effective neck structure called FBLNeck to balance these two losses for optimization. Although the BNNeck proposed in \cite{luo2019bag} can balance these two losses to achieve training convergence, it cannot guarantee optimal results. We first introduce BNNeck into our model, and the result can be seen in the first row in Table \ref{tb:fblneck}. The use of BLNeck improves the rank-1/mAP by 2.1\%/3.5\% over the use of BNNeck. Obviously, our BLNeck can better balance the effects of the triplet loss and softmax loss. We then further combine BLNeck and FNeck as FBLNeck, which can extract both local and global features. As shown in the third row in Table \ref{tb:fblneck}, both the rank-1 accuracy and mAP are greatly improved, which demonstrates that the fine-grained information extracted from shallower depths by FNeck is helpful. It is worth noting that our FBLNeck can be removed at inference times, thus making the model more lightweight and efficient. Experiments about introducing FBLNeck into OSNet are shown in the supplementary materials.}
\begin{table}[h]
	\begin{center}
		\begin{tabular}{l|c|c|c}
			\hline 
			Architecture & Feature-dim & rank-1 & mAP \\
			\hline
			+BNNeck & 512 & 91.0 & 77.6 \\
			+BLNeck & 512 & 93.1 & 81.1 \\
			+FBLNeck & 768 & \textbf{93.7} & \textbf{83.7} \\
			\hline
		\end{tabular}
	\end{center}
	\caption{Effect of each component of FBLNeck. The backbone is the body of CDNet. All experiments are conducted on Market1501.}
	\label{tb:fblneck}
\end{table}
\begin{table}[ht]
	\begin{center}
		\begin{tabular}{l|c|c|c}
			\hline
			Network & Param(M) & rank-1 & mAP \\
			\hline
			CDNet & 1.80 & \textbf{93.7} & \textbf{83.7} \\
			CDNet\_std & 1.79 & 91.8 & 79.7 \\
			CDNet\_v35 & 1.61 & 93.0 & 82.0 \\
			\hline 
		\end{tabular}
	\end{center}
	\caption{The effect of combined patterns and search. std denotes that the CBlock is changed to a single branch. v35 indicates that the kernel combination is fixed to (3,5) with the same number of layers as in CDNet. All experiments are conducted on Market1501.}

	\label{tb:effect_cs}
\end{table}
\vspace{-0.6cm}
\paragraph{- Effect of combination and search.} To verify the effect of the kernel combination pattern, we change the structure of CBlock to a single branch with the same number of building blocks; thus, the constructed network is called CDNet\_std. As shown in Table \ref{tb:effect_cs}, compared with those of the original CDNet, both rank-1 and mAP of CDNet\_std drop dramatically, indicating that our combined pattern learning is much effective. Both CDNet and CDNet\_v35 share the same number of blocks in each stage, but the latter fixes the kernel combination to 3$\times$3 and 5$\times$5 in each CBlock. Ultimately, the latter suffers from its lack of combination variety, resulting in suboptimal performance.

\section{Conclusion}
\cgreen{In this paper, we introduce the Combined Depth Space, and obtain a lightweight and efficient network called CDNet via top-2 sample search, which is effectively for ReID. Our experiments show that the proposed Fine-grained Balance Neck effectively balances the effects of triplet loss and softmax loss. The extensive experiments also further demonstrate that CDNet outperforms state-of-the-art lightweight networks proposed for person re-identification task.}

\section{Acknowledgements}
This work was supported partially by the NSFC(U1911401,U1811461), Guangdong NSF Project (No. 2020B1515120085, 2018B030312002), Guangzhou Research Project (201902010037), and Research Projects of Zhejiang Lab (No. 2019KD0AB03), and the Key-Area Research and Development Program of Guangzhou (202007030004).

\section{Supplementary Material}
\subsection{Details for Top-k Sample Search}
Different from normal NAS, we aim to search for most suitable network architectures for ReID. Therefore, we directly search the network architectures on Market1501 \cite{zheng2015scalable}. In Market1501, the dataset is divided into a training set with 12936 images of 751 persons and a testing set of 750 persons containing 3368 query images and 19732 gallery images. According to the proposed Top-k Sample Search, we need to split the training set into a new training set and a validation set. For each identity, we select 4 images to construct the validation set. We utilize the triplet sampler as in \cite{luo2019bag} on both training set and validation set to prepare the batch data. Particularly, we choose the batch size as 64 and the number of identities per batch is 16, namely each identity has 4 instances per batch. SGD optimizer with weight decay 3e-4 and Adam optimizer with weight decay 0.001 are utilized for the network parameters and the architecture parameters optimization respectively. The network parameters learning rate $\eta_w$ starts from 0.025 and decays to 0.0001 via a cosine lr\_scheduler while architecture parameters learning rate $\eta_\alpha$ starts from 3e-4 and decayed by 0.1 at 80, 160 epochs. Totally, we train the search network for 240 epochs. During the training process, softmax loss and triplet loss are jointly utilized for optimization and the margin of triplet loss is 0.3.

In Tab. \ref{tb:all_archs}, we show the inner structures of CNet and CDNet via Top-k Sample Search, where $k \in \{1, 2, 3, 4 \}$. As analyzed in main manuscript (section 4.1), the architectures searched via top-2 sample search almost achieve the best performance. As for CNet, the best architecture (top-2 CNet) tends to select large kernel combination thus makes up the defect of shallow depth. Note that the depth of CDNet can vary from 6 to 12 and none of the searched architectures choose the biggest depth. As we can see, the best architecture of CDNet(top-2 CDNet) just add one more CBlock at each stage. Since top-2 CDNet has enough depth, its kernels size tend to be smaller compared with top-2 CNet. Therefore, it is inappropriate to randomly add the depth of the network.
\begin{table*}[h]
    \begin{center}
         \begin{tabular}{c|c|c|c|c|c|c|c|c}
         \hline
         \multirow{2}{*}{Layer} & \multicolumn{4}{c|}{CNet($k_1, k_2$)} & \multicolumn{4}{c}{CDNet($k_1, k_2, r$)} \\
         \cline{2-9}
         & top-1 & top-2 & top-3 & top-4 & top-1 & top-2 & top-3 & top-4 \\
         \hline 
         1 & (3,5) & (5,7) & (5,7) & (5,7) & (3,7,2) & (3,5,1) & (5,7,2) & (3,9,2) \\
         2 & (7,9) & (7,9) & (3,9) & (3,5) & (3,9,1) & (3,7,2) & (5,7,2) & (5,9,1) \\
         3 & (3,7) & (7,9) & (3,5) & (5,7) & (5,9,2) & (5,7,2) & (3,5,2) & (5,7,2) \\
         4 & (3,9) & (7,9) & (3,5) & (3,7) & (3,7,1) & (5,9,1) & (5,7,1) & (3,7,1) \\ 
         5 & (5,9) & (7,9) & (3,5) & (3,9) & (3,9,2) & (5,7,2) & (7,9,2) & (5,7,1) \\
         6 & (7,9) & (3,5) & (5,7) & (3,7) & (7,9,2) & (5,7,1) & (3,7,1) & (5,7,1) \\
         \hline
         Depth & 6 & 6 & 6 & 6 & 10 & 9 & 10 & 8 \\ 
         \hline
        \end{tabular}
    \end{center}
    \caption{The inner structures of all searched architectures. $k_1, k_2$ denote the kernel size of two branches in CBlock respectively. $r$ denotes the repeated times of CBlock for CDBlock }
    \label{tb:all_archs}
\end{table*}
\subsection{Visualization for the effect of BLNeck}
\begin{figure}[h]
	\begin{center}
		\includegraphics[width=0.8\linewidth]{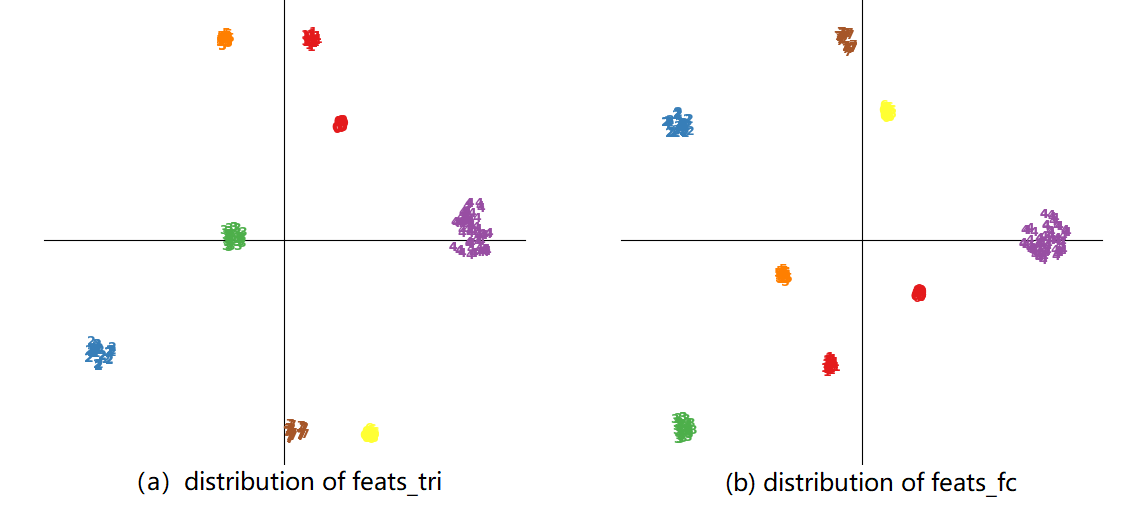}
	\end{center}
	\caption{We first select 8 identities and use t-sne method to project their features into 2-dimension space. The feats\_tri constrained by triplet loss are on the left side and the feats\_fc constrained by softmax loss are on the right side.}
	\label{fig:fbl_tf8}
\end{figure}
\begin{figure}[h]
	\begin{center}
		\includegraphics[width=0.8\linewidth]{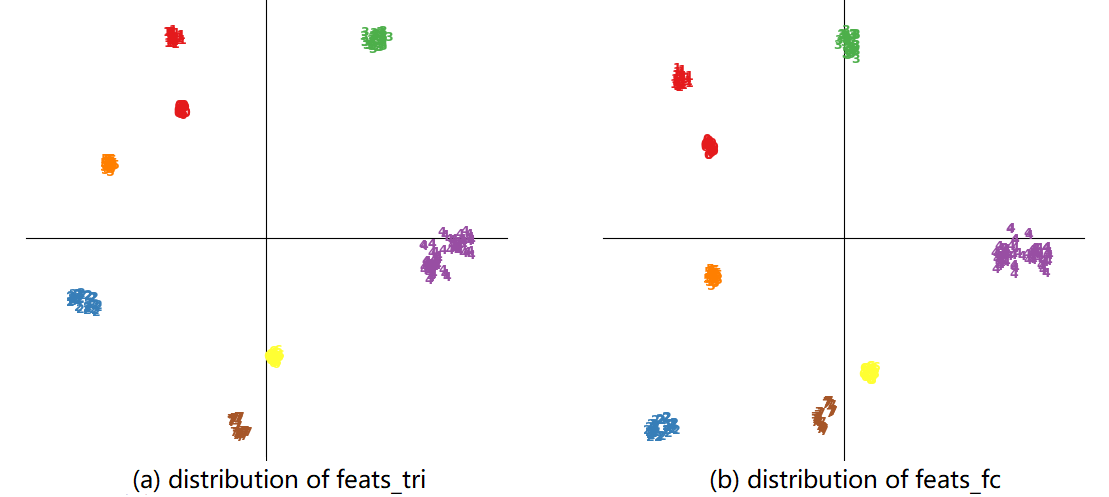}
	\end{center}
	\caption{The feats\_tri constrained by triplet loss are on the left side and the feats\_fc constrained by softmax loss are on the right side.}
	\label{fig:bn_tf8}
\end{figure}
To further understand the influence of BLNeck, we visualize the distribution of the features which are processed before and after by BLNeck. As shown in Fig. \ref{fig:fbl_tf8}, BLNeck learns to map the feats\_tri to another embedding space which is fitter to softmax loss. According to the distribution of feats\_fc, there are clear angle margins between each identity. BNNeck proposed in \cite{luo2019bag} can easily balance the constraint of triplet loss and triplet loss. In Fig. \ref{fig:bn_tf8}, we show that BNNeck only makes a little adjustment of the features, and they still affect greatly by each other, thus the training can not converge peacefully. Compared with BNNeck, the proposed BLNeck has stronger ability to learn the mapping from embedding space constrained by triplet loss to that constrained by softmax loss.
\begin{table}[h]
	\begin{center}
		\begin{tabular}{l|c|c|c}
			\hline 
			Architecture & Param(M) & rank-1 & mAP \\
			\hline
			+Softmax & 2.2 & 92.5 & 80.0 \\
			+Softmax\_Triplet & 2.2 & 93.1 & 82.2 \\
			+FBLNeck & 1.7 & \textbf{93.6} & \textbf{83.4} \\
			\hline
		\end{tabular}
	\end{center}
	\caption{Effect of  FBLNeck. The backbone is the body of OSNet. +Softmax denotes OSNet is trained with only softmax loss. +Softmax\_Triplet denotes that OSNet is trained with softmax and triplet loss. +FBLNeck denotes that OSNet is trained with FBLNeck as its attached head. All experiments are conducted by us on Market1501.}
	\label{tb:fblneck_OSNet}
\end{table}
\begin{table*}[h]
    \begin{center}
         \begin{tabular}{c|c|c|c|c|c|c|c|c|c}
         \hline
         \multirow{2}{*}{$\beta$} & \multirow{2}{*}{$\gamma$} & \multicolumn{4}{c|}{CDNet} & \multicolumn{4}{c}{OSNet} \\
         \cline{3-10}
         & & Param(M) & FLOPS(M) & rank-1 & mAP & Param(M) & FLOPs(M) & rank-1 & mAP \\
         \hline
         1.0 & 1.0 & 1.8 & 948.8 & \textbf{95.1} & \textbf{86.0} & 2.2 & 978.9 & 94.8 & 84.9 \\
         1.0 & 0.75 & 1.8 & 533.7 & \textbf{94.7} & \textbf{85.0} & 2.2 & 550.7 & 94.4 & 83.7 \\
         1.0 & 0.5 & 1.8 & 237.3 & \textbf{93.3} & \textbf{82.3} & 2.2 & 244.9 & 92.0 & 80.3 \\
         1.0 & 0.25 & 1.8 & 59.4 & \textbf{86.9} & \textbf{69.5} & 2.2 & 61.5 & \textbf{86.9} & 67.3 \\
         \hline 
         0.75 & 1.0 & 1.0 & 552.3 & \textbf{94.7} & \textbf{85.1} & 1.3 & 571.8 & 94.5 & 84.1 \\
         0.75 & 0.75 & 1.0 & 310.7 & 94.2 & \textbf{84.3} & 1.3 & 321.7 & \textbf{94.3} & 82.4 \\
         0.75 & 0.5 & 1.0 & 138.1 & \textbf{93.1} &\textbf{ 81.4} & 1.3 & 143.1 & 92.9 & 79.5 \\
         0.75 & 0.25 & 1.0 & 34.6 & \textbf{86.8} & \textbf{69.3} & 1.3 & 35.9 & 85.4 & 65.5 \\
         \hline 
         0.5 & 1.0 & 0.5 & 262.0 & \textbf{93.4} & \textbf{83.8} & 0.6 & 272.9 & \textbf{93.4} & 82.6 \\
         0.5 & 0.75 & 0.5 & 147.4 & \textbf{93.9} & \textbf{83.5} & 0.6 & 153.6 & 92.9 & 80.8 \\ 
         0.5 & 0.5 & 0.5 & 65.5 & \textbf{92.5} & \textbf{80.3} & 0.6 & 68.3 & 91.7 & 78.5 \\
         0.5 & 0.25 & 0.5 & 16.4 & 84.3 & \textbf{66.4} & 0.6 & 17.2 & \textbf{85.4} & 66.0 \\
         \hline 
         0.25 & 1.0 & 0.1 & 77.9 & 91.7 &\textbf{ 79.7} & 0.2 & 82.3 & \textbf{92.2} & 77.8 \\
         0.25 & 0.75 & 0.1 & 43.8 & \textbf{91.8} & \textbf{78.8} & 0.2 & 46.3 & 91.6 & 76.1 \\
         0.25 & 0.5 & 0.1 & 19.5 & \textbf{89.3} & \textbf{75.0} & 0.2 & 20.6 & 88.7 & 71.8 \\
         0.25 & 0.25 & 0.1 & 4.88 & \textbf{79.6} & \textbf{59.4} & 0.2 & 5.2 & 79.1 & 56.0  \\
         \hline 
    \end{tabular}
    \end{center}
    \caption{Results(\%) of varying width multiplier $\beta$ and resolution multiplier $\gamma$ for CDNet and OSNet. The resolution is 256 $\times$ 128, 192 $\times$ 96, 128 $\times$ 64 and 64 $\times$ 32 for $\gamma = 1.0, \gamma = 0.75, \gamma = 0.5$ and $\gamma = 0.25$ respectively.}
    \label{tb:scaling}
\end{table*}
\subsection{Scaling CDNet with width multiplier and resolution multiplier}
As shown in Tab. \ref{tb:scaling}, we scale the CDNet for specific devices with limited computational resources via adjusting the width multiplier $\beta$ and resolution multiplier $\gamma$. When fixing the $\beta$ and shrinking the $\gamma$ from 1.0 to 0.5, the number of FLOPs decreases significantly while the rank-1 just drops smoothly. However, we find that both rank-1 and mAP drop dramatically when $\beta$ decreases to 0.25. This is because the resolution of images are reduced to 4$\times$2 at stage 3, which is too small to learn effective information via convolution at stage 3. It is worth noting that CDNet still can achieve 91.7\%/79.7\% at rank-1/mAP when both $\beta$ and  $\gamma$ are set to 0.25 and 1.0 respectively (with merely 0.1M parameters and 77.9M FLOPs). This suggests that CDNet has great potential for deployment in edge devices such as surveillance cameras with limited computational resources. On the right side, we also report the result of OSNet accordingly. Apparently, CDNet outperforms OSNet at rank-1 and mAP for most settings with lower parameters and FLOPs, which demonstrates the robustness of combined pattern learning.

\subsection{Implementation of FBLNeck in OSNet}
As analyzed in main manuscript (section 4.5), FBLNeck could take advantage of the combination of triplet loss and softmax loss and utilize fine-grained information, which leads to high performance. As a newly proposed neck, FBLNeck can be inserted to other models easily in addition to CNet and CDNet. Here we investigate whether inserting FBLNeck to OSNet could enhance the performance of OSNet \cite{zhou2019omni}. As shown in Tab. \ref{tb:fblneck_OSNet}, with FBLNeck implemented with OSNet, both rank-1 accuracy and mAP increase, especially for mAP. It is fair to say that the proposed FBLNeck could better balance the effect of triplet loss and softmax loss to help models achieve higher performance. Note that the number of parameters of OSNet+FBLNeck is reduced to 1.7M, since we remove the attached head of OSNet and our FBLNeck is removable at inference time. Moreover, comparing OSNet + FBLNeck and CDNet, our searched CDNet outperforms OSNet with less parameters, which means that our proposed search algorithm can obtain models that are computationally efficient and suitable for ReID.
\begin{table}[h]
	\begin{center}
		\begin{tabular}{l|c|c|c}
			\hline
			Model & Param(M) & FLOPs(M) & Top-1 \\
			\hline 
			CARS\cite{yang2020cars} & 5.1  & 519 & 75.2 \\
			FBNet\cite{wu2019fbnet} & 5.5  & 375 & 74.9 \\
			GDAS\cite{dong2019searching} & 5.3  & 581 & 74.0 \\
			DARTS\cite{liu2018darts} & 4.7  & 574 & 73.3 \\
			OSNet\cite{zhou2019omni} & 2.7  & 1511 & \textbf{75.5} \\
			GhostNet\cite{han2020ghostnet} & 5.2 & 141 & 73.9 \\
			MobileNetV2\cite{sandler2018mobilenetv2} & 3.4  & 300 & 73.0 \\
			MobileNetV3\cite{howard2019searching} & 5.4  & 219 & 75.2\\
			\textbf{CDNet(ours)} & \textbf{2.5} & 1571 & 75.1\\
			\hline
		\end{tabular}
	\end{center}
	\caption{Top-1(\%) accuracy on ImageNet-2012 validation set.}
	\label{tb:imagenet}
\end{table}

\subsection{Evaluation on ImageNet}
In this section, we evaluate the transferability of proposed CDNet on the ImageNet \cite{russakovsky2015imagenet}. The size of image is resized to $224 \times 224$. Random horizontal flip and random crop are utilized for data augmentation. We adopt the training scheme as in \cite{liu2018darts}. CDNet is trained for 240 epochs with weight decay 3e-5 and initial SGD learning rate 0.1(decayed by a factor of 0.97 after each epoch). As shown in Tab. \ref{tb:imagenet}, although our CDNet is originally designed for ReID, it still achieves comparable performance among lightweight networks which are specially designed for classification. In particular, CDNet outperforms MobileNetV2 by 2.1\% with fewer number of parameters. It is worth noting that CDNet surpasses GDAS and DARTS by 1.1\% and 1.8\% respectively, which indicates that the proposed CDS also has great potential for classification tasks. Obviously, the superior performance on classification tasks demonstrates the benefit of learning combined pattern information. 

\subsection{Inference time on Market1501}
As shown in Tab. \ref{tb:time}, with about 3$\times$ fewer FLOPs and 13$\times$ fewer parameters, CDNet achieves competitive performance with lower latency compared with BagofTrick, which is representative of those models utilizing ResNet as backbone. Besides, compared with the other two lightweight models, CDNet achieves the best performance with faster speed.
\begin{table}[h]\scriptsize
    \begin{center}
         \begin{tabular}{c|c|c|c|c|c}
         \hline
         Model& Param(M) & Flops(M) & Times(s) & Rank-1 & mAP  \\
         \hline
         BagofTrick \cite{luo2019bag}& 25.1 & 4053.3 & 248.3 & 94.5 & 85.9 \\
         OSNet\cite{zhou2019omni} & 2.2 & 979.0 & 156.0 & 94.8 & 84.9 \\
         GDAS\cite{dong2019searching} & 4.0 & 1109.2 & 150.9 & 89.1 & 73.2 \\
         CDNet(ours) & 1.8 &  955.1 & 142.4 & 95.1 & 86.0 \\
         \hline
    \end{tabular}
    \end{center}
    \caption{All experiments are conducted on Market1501 with single RTX 2080 and the batch size is 128. The time is the average of 5 times inference time on test set with 19281 images.}
    \label{tb:time}
\end{table}

{\small
\bibliographystyle{ieee_fullname}
\bibliography{egbib}
}

\end{document}